\documentclass[conference]{IEEEtran}
\IEEEoverridecommandlockouts    % to override locked commands

\usepackage{multirow}
\usepackage{lipsum}
\usepackage{amsmath}
\usepackage{amssymb}
\usepackage[ruled,vlined]{algorithm2e}
\usepackage{graphicx}
\usepackage{booktabs}
\usepackage{placeins}
\usepackage{array}
\usepackage{tabularx}
\usepackage{diagbox}
\usepackage{makecell}
\usepackage[switch]{lineno}

\graphicspath{{./Figures/}}
\usepackage[utf8]{inputenc}

\title{\vspace*{0.1in}Towards Reliable Zero-Shot Crowd Forecasting: Evaluating Time Series Foundation Models for Special Event Pedestrian Forecasting}

\author{
Ziteng Li$^{1}$,
Yanan Xin$^{1}$,
Tina Comes$^{2,3}$,
Serge Hoogendoorn$^{1}$
\thanks{$^{1}$Delft University of Technology (TU Delft), Department of Transport \& Planning, Delft, The Netherlands.}
\thanks{$^{2}$Delft University of Technology (TU Delft), Technology, Policy and Management, Delft, The Netherlands.}
\thanks{$^{3}$German Aerospace Center (DLR), Institute for the Protection of Terrestrial Infrastructures, Sankt Augustin, Germany.}
\thanks{Corresponding author: Ziteng Li. Email: \texttt{Z.Li-27@tudelft.nl}.}
}

\begin{document}
% \linenumbers

\maketitle 
\thispagestyle{empty}
\pagestyle{empty}

%%%%%%%%%%%%%%%%%%%%%%%%%%%%%%%%%%%%%%%%%%%%%%%%%%%%%%%%%%%%%%%%%%
\begin{abstract}
Managing massive crowds during infrequent special events requires reliable real-time pedestrian-flow forecasting to ensure public safety and operational efficiency. However, supervised forecasting methods face limitations in these contexts due to scarce historical data, heterogeneous data distributions, and short in-event observation windows. To effectively support operational decision-making, forecasts should provide not only accurate point estimates but also informative predictive uncertainty. Probabilistic uncertainty quantification plays a critical role in this aspect, particularly capturing sudden volatility and tail risks. This paper investigates pretrained time series foundation models as a lightweight approach for zero-shot probabilistic forecasting without extensive local retraining. Using decision-oriented metrics tailored to short events, we conduct a comprehensive assessment of two time series foundation models on crowd forecasting, with the SAIL2025 event as a use case. We then distill practical insights for crowd managers, specifying when zero-shot forecasts remain operationally reliable.
\end{abstract}

%%%%%%%%%%%%%%%%%%%%%%%%%%%%%%%%%%%%%%%%%%%%%%%%%%%%%%%%%%%%%%%%%%
\section{Introduction}
\label{sec:introduction}
Amsterdam hosts the SAIL maritime festival every five years. 2025 marked the 50th year celebration of SAIL as well as the 750th anniversary of Amsterdam. This 5-day maritime event is the largest in the world, drawing roughly 2.5 million people to the center of Amsterdam. Managing such large crowds in a busy city center requires real time insights and reliable forecasts of pedestrian flows to support proactive infrastructure adjustments, staffing and routing decisions, and emergency evacuation planning \cite{comes2024ai, yuan2016comparison}. In this decision setting, reliable forecasts demand probabilistic uncertainty quantification calibrated for coverage and informative about dispersion \cite{gneiting2007strictly} besides accurate point predictions \cite{gneiting2014probabilistic}. In addition, reliable forecasts require robustness to incomplete observations, disturbances, and distributional shifts \cite{yoon2022robust}.

While crowd forecasting has moved from simulation-based approaches to supervised data-driven models trained on local historical data \cite{lv2014traffic}, this paradigm encounters a technical limitation in the context of infrequent events such as SAIL2025. Modern supervised algorithms, such as those utilizing recurrent architectures or spatio-temporal graph neural networks, learn temporal dependencies and spatio-temporal structure from extensive historical sequences \cite{wen2025dynamic}. However, the SAIL event was not held in 2020 due to the COVID-19 pandemic, which means historical event data is effectively unavailable for supervised training in a consistent way. Moreover, the observation window available for this study is only five days, which limits the ability of conventional supervised models to learn stable periodic structure and to generalize under rolling multi-step forecasting. % In addition, in such an event, crowd volumes can rise sharply within minutes, and local congestion may emerge from small disruptions such as temporary blockages, service problems, or control measures that reroute pedestrians \cite{hoogendoorn2025pedestrian}. For crowd managers, the challenge is not only fast forecasting under time pressure, but also providing useful information that reflects a highly dynamic and uncertain situation and supports trend-aware adaptation \cite{comes2020coordination}.

In recent years, advances in large language models and foundation models have led to pretrained time series foundation models. Unlike task-specific supervised models that rely on substantial local historical data, foundation models are pretrained on large and heterogeneous time series corpora and support direct deployment on new series in a zero-shot setting \cite{das2024decoder,ansari2025chronos}. This property is attractive for short-term special-event operations, where time-consuming model development and data-intensive retraining are impractical \cite{liu2024frequency}. In addition, recent foundation models can produce probabilistic forecasts, for example predictive quantiles such as the 10th, 50th, and 90th percentiles \cite{ansari2025chronos}. This output matches operational needs, since many decisions depend on plausible ranges and risk-aware planning rather than a single point estimate \cite{boersma2017communicating}. 

In this paper, we propose a systematic evaluation of time series foundation models for rolling multi-step pedestrian flow forecasting, applied to the SAIL2025 event, with a focus on operationally reliable predictions for crowd-management decision support. Our main contributions are:

\begin{itemize}
    \item \textbf{Decision-oriented reliability evaluation metrics for short events:} We design a set of operational reliability evaluation metrics for rolling multi-step probabilistic forecasting in short-duration special events, combining uncertainty estimates with temporal stability and forecasting accuracy. These metrics enable quantitative model comparison and reliability assessment for crowd managers.
    \item \textbf{Systematic zero-shot assessment of foundation models:} We conduct a systematic zero-shot evaluation of pretrained time series foundation models in the proposed setting, comparing their performance across lead times, different data distributions, and missing data case.
    %\item \textbf{Empirical insights for crowd management:} We distill practical findings for crowd managers on when zero-shot foundation models provide usable forecasts, where they break down under high volatility and limited observations, and how to interpret their uncertainty outputs for risk-aware operational decisions.
\end{itemize}

\section{Related work}
Deep learning has become the dominant method in time series forecasting. Recurrent neural networks such as DeepAR \cite{salinas2020deepar}, residual architectures like N-BEATS \cite{oreshkin2019n}, and transformer-based models \cite{zeng2023transformers} have demonstrated strong performance across diverse forecasting tasks. However, these methods generally require task-specific training and sufficient historical data \cite{li2024deep}, which limits their applicability in data sparsity situations.

Recently, pretrained time series foundation models have attracted growing interest. After pretraining, the model can be directly applied to unseen datasets in a zero-shot setting or adapted with limited fine-tuning. Representative examples include TimesFM \cite{das2024decoder} and Chronos \cite{ansari2025chronos}. 

Several deep learning methods also focus on estimating uncertainty within time-series forecasting. Quantile-based forecasting models estimate upper and lower quantiles to construct prediction intervals, rather than reporting only point forecasts \cite{shao2025quantileformer}. Recent studies propose prediction for time series to obtain prediction intervals with stated coverage under temporal change and non-stationarity \cite{sunconformal,li2025neural}. Time-series foundation models follow the same probabilistic perspective: Chronos provides pretrained probabilistic forecasting, while TimesFM~2.5 supports quantile-based outputs for uncertainty estimation.

\section{Experiments}
\subsection{Data}

The crowd count data used in this study were collected from 11 pedestrian flow count sensors deployed by the Amsterdam municipality as part of Amsterdam’s Crowd Monitoring System. The sensors were installed along pedestrian walkways to monitor the main pedestrian traffic flow. Each sensor independently monitors two opposite directions (180 degrees apart), which results in 22 independent time series. For the SAIL event, this pattern typically follows a daily cycle: pedestrian volumes increase in the morning, remain high from daytime to early evening, and then gradually decline late at night. Each record contains a sensor identifier, direction, a timestamp, and a pedestrian count. The counts are aggregated into contiguous 90-second intervals for each sensor–direction pair. To ensure data integrity, negative counts are clipped to zero and missing values are imputed with zero. Missing observations are treated as potential data quality artifacts, and their impact is analyzed in the missing data sensitivity case study.

This section introduces two state-of-the-art pretrained time series foundation models evaluated in this study, TimesFM and Chronos-2. 

\paragraph{TimesFM  \cite{das2024decoder}}
TimesFM is a time series foundation model developed by Google for zero-shot forecasting. TimesFM uses a decoder-only Transformer and a patch-based representation that groups consecutive values into segments, which improves efficiency and supports long-context modeling. Pretraining uses a large corpus of real-world time series at web scale, and the training data also includes synthetic time series. This study uses TimesFM 2.5 with a 200M-parameter configuration for multi-step forecasting.

\paragraph{Chronos-2  \cite{ansari2025chronos}}
Chronos, developed by Amazon, is a family of pretrained forecasting models that frames time series forecasting as a language modeling problem. Chronos uses a Transformer architecture based on T5 \cite{raffel2020exploring}, which is a series of large language models developed by Google AI introduced in 2019. The method converts real-valued time series into discrete tokens through scaling and quantization, and the model forecasts by predicting future tokens. Pretraining uses a large collection of publicly available time series datasets, and it adds synthetic series generated from Gaussian processes to improve generalization. In this paper, we use Chronos-2, which extends Chronos with updated training and improved forecasting capability across broader task settings.

\subsection{Experimental Design}

This study conducts multi-step pedestrian flow count forecasting with a one-hour forecast horizon, a 90-second sampling interval, and a rolling prediction with a stride of one time step. The lead time index $t$ ranges from 1 to 40, where $t=40$ corresponds to one hour ahead. The index $i$ denotes a target timestamp, and the realized flow count at lead time $t$ is $y_{i,t}$. 

The probabilistic forecast models output a lower bound $\hat{\ell}_{i,t}$ and an upper bound $\hat{u}_{i,t}$ for target $i$ at lead time $t$. We select the 10th and 90th percentile predictions as bounds, so $[\hat{\ell}_{i,t}, \hat{u}_{i,t}]$ forms a central 80\% prediction interval. We evaluate interval coverage of the realized observation across instances and lead times. The 10th–90th percentile interval is commonly used for uncertainty visualization and assessment in time-series foundation model forecasting \cite{gruver2023large}, as well as in analyses of traffic and public-transit ridership during special events to characterize variability and reliability \cite{yildirimoglu2013experienced,santanam2024public}.

We also compare the models with different input lengths (i.e., context windows) using two strategies: fixed context and increasing context. The fixed strategy uses the most recent 960 ground-truth values (24h) to predict the next 40 steps. The increasing strategy appends each newly observed ground-truth value, increasing the input length over days. Neither model truncates the input because its maximum allowable context length exceeds the five-day length. Since Chronos-2 supports multivariate forecasting, we also evaluate Chronos-2 variants with additional time features (date and hour) as input covariates under both fixed and increasing strategies. 

All experiments were run on an HP laptop equipped with an Intel Core i7-1355U CPU (13th Gen) with 12 logical cores (x86\_64), and 16 GB RAM. No GPU acceleration was used. In terms of computational feasibility, all models required less than 0.5 seconds to generate a single 40-step prediction, substantially below the 90-second data update frequency.

\subsection{Evaluation Metrics}

We design a set of metrics to evaluate the operational reliability of model outputs with regard to the uncertainty estimates. In this paper, reliability covers three aspects: (i) predictive accuracy, (ii) temporal stability, and (iii) the quality of predictive uncertainty. These metrics provide crowd managers with quantitative indicators to compare models and to assess the reliability of forecasts. 

Adopting the \emph{Prediction-Interval Coverage Probability (PICP)} metric~\cite{khosravi2011comprehensive},  we extend the coverage indicator $c_i$ to $c_{i,t}$ for multi-step forecasting, indicating whether the prediction interval covers the ground-truth $y_{i, t}$.
\begin{equation}
c_{i,t}
=
\begin{cases}
1, & \text{if } \hat{\ell}_{i,t} \le y_{i,t} \le \hat{u}_{i,t},\\
0, & \text{otherwise},
\end{cases}
\quad
t \in \{1,\dots,40\}.
\end{equation}
All metrics below are defined based on this indicator.

\subsubsection{Predictive Accuracy Metrics}

\paragraph{Full-Horizon Coverage (FHC)}
Full-Horizon Coverage evaluates the reliability of an entire multi-step forecast trajectory. A forecast round is considered reliable only if the predicted interval covers the ground truth at all lead 40 time steps. Formally,
\begin{equation}
\text{FHC}
=
\frac{1}{N}
\sum_{i=1}^{N}
\prod_{t=1}^{40} c_{i,t},
\end{equation}
where $N$ denotes the total number of target forecasts. The product $\prod_{t=1}^{40} c_{i,t}$ equals $1$ if and only if all 40 step predictions fall within the interval; otherwise, it equals $0$. Therefore, $\text{FHC}$ measures the proportion of forecasts in which the 40-step prediction intervals fully contain ground truth values.

\paragraph{Nearest-Horizon Coverage (NHC)}
In contrast to FHC, {\em Nearest-Horizon Coverage} focuses only on one-step-ahead ($t=1$) forecast reliability. 
\begin{equation}
\text{NHC}
=
\frac{1}{N}
\sum_{i=1}^{N}
c_{i,1}.
\end{equation}
NHC reports the proportion of forecast instances for which the interval covers the ground truth at the nearest forecasting step, regardless of whether misses occur at later time steps.

\begin{table*}[htbp]
  \centering
  \small
  \setlength{\tabcolsep}{1.5pt}
  \caption{Core Operational Hours \textbf{08:00-22:00}. Format: \textbf{FHC\,/\,NHC\,/\,SLT}. Metrics are ranked individually (Bold: 1st, Underline: 2nd, SLT is ceiled).}

  \renewcommand{\arraystretch}{1.3}
  \newcolumntype{C}[1]{>{\centering\arraybackslash}m{#1}}
  % \begin{tabularx}{\textwidth}{|m{3.5cm}|*{6}{>{\centering\arraybackslash}X|}}
  \label{tab:over_all}
  \begin{tabularx}{\textwidth}{|C{3.2cm}|*{6}{>{\centering\arraybackslash}X|}}

  \hline

  \multicolumn{1}{|c|}{\diagbox[width=3.2cm,height=8.0ex,dir=SW]{\textbf{Sensor ID}}{\textbf{FHC/NHC/SLT}}} &
  \multicolumn{1}{c|}{\makecell[c]{\textbf{TimesFM}\\\textbf{(Fixed)}}} &
  \multicolumn{1}{c|}{\makecell[c]{\textbf{TimesFM}\\\textbf{(Increasing)}}} &
  \multicolumn{1}{c|}{\makecell[c]{\textbf{Chronos-2}\\\textbf{(Fixed)}}} &
  \multicolumn{1}{c|}{\makecell[c]{\textbf{Chronos-2}\\\textbf{(Increasing)}}} &
  \multicolumn{1}{c|}{\makecell[c]{\textbf{Chronos-2}\\\textbf{(Fixed}\\\textbf{+ TimeFeature)}}} &
  \multicolumn{1}{c|}{\makecell[c]{\textbf{Chronos-2}\\\textbf{(Increasing}\\\textbf{+ TimeFeature)}}} \\
  \hline

  GASA-01-A1\_135 & \underline{0.558}\,/\,0.777\,/\,25 & 0.547\,/\,0.776\,/\,25 & 0.505\,/\,0.748\,/\,24 & 0.529\,/\,0.738\,/\,24 & 0.557\,/\,\textbf{0.796}\,/\,\underline{26} & \textbf{0.575}\,/\,\underline{0.794}\,/\,\textbf{27} \\
  
  GASA-01-A1\_315 & \underline{0.556}\,/\,0.776\,/\,26 & 0.543\,/\,0.775\,/\,25 & 0.482\,/\,0.755\,/\,23 & 0.534\,/\,0.754\,/\,24 & 0.545\,/\,\underline{0.802}\,/\,\underline{27} & \textbf{0.583}\,/\,\textbf{0.804}\,/\,\textbf{28} \\
  GASA-01-A2\_135 & 0.575\,/\,0.767\,/\,26 & 0.560\,/\,0.761\,/\,26 & 0.542\,/\,0.752\,/\,25 & \underline{0.606}\,/\,0.782\,/\,\underline{27} & 0.572\,/\,\underline{0.800}\,/\,\underline{27} & \textbf{0.638}\,/\,\textbf{0.808}\,/\,\textbf{28} \\
  GASA-01-A2\_315 & 0.564\,/\,0.784\,/\,26 & 0.565\,/\,\underline{0.787}\,/\,26 & 0.530\,/\,0.757\,/\,25 & \underline{0.571}\,/\,0.768\,/\,26 & 0.550\,/\,\textbf{0.799}\,/\,\underline{27} & \textbf{0.619}\,/\,\textbf{0.817}\,/\,\textbf{28} \\
  
  GASA-01-B\_135 & \underline{0.632}\,/\,\underline{0.779}\,/\,\textbf{29} & 0.625\,/\,0.778\,/\,\underline{28} & 0.592\,/\,0.754\,/\,26 & 0.624\,/\,0.765\,/\,27 & 0.595\,/\,0.778\,/\,27 & \textbf{0.649}\,/\,\textbf{0.782}\,/\,\textbf{29} \\
  GASA-01-B\_315 & 0.583\,/\,\underline{0.788}\,/\,27 & 0.575\,/\,0.786\,/\,27 & 0.562\,/\,0.766\,/\,26 & \underline{0.596}\,/\,0.786\,/\,27 & 0.575\,/\,\textbf{0.808}\,/\,\underline{28} & \textbf{0.628}\,/\,0.802\,/\,\textbf{29} \\
  GASA-01-C\_135 & 0.622\,/\,0.763\,/\,26 & 0.619\,/\,0.758\,/\,26 & 0.604\,/\,0.763\,/\,26 & \underline{0.635}\,/\,0.768\,/\,\underline{27} & 0.629\,/\,\underline{0.774}\,/\,\underline{27} & \textbf{0.650}\,/\,\textbf{0.779}\,/\,\textbf{28} \\
  GASA-01-C\_315 & 0.617\,/\,\underline{0.765}\,/\,\underline{27} & 0.612\,/\,0.753\,/\,\underline{27} & 0.574\,/\,0.723\,/\,25 & \underline{0.625}\,/\,0.756\,/\,\underline{27} & 0.587\,/\,0.758\,/\,26 & \textbf{0.634}\,/\,\textbf{0.765}\,/\,\textbf{28} \\
  \hline

  GASA-02-01\_135 & \underline{0.549}\,/\,0.791\,/\,26 & 0.539\,/\,0.787\,/\,26 & 0.485\,/\,0.762\,/\,24 & 0.520\,/\,0.773\,/\,25 & 0.544\,/\,\underline{0.798}\,/\,\underline{26} & \textbf{0.571}\,/\,\textbf{0.799}\,/\,\textbf{27} \\
  GASA-02-01\_315 & \textbf{0.564}\,/\,\underline{0.817}\,/\,\underline{27} & \underline{0.546}\,/\,0.814\,/\,\underline{27} & 0.428\,/\,0.778\,/\,23 & 0.451\,/\,0.764\,/\,24 & 0.475\,/\,0.815\,/\,25 & 0.541\,/\,\textbf{0.827}\,/\,\textbf{27} \\
  GASA-02-02\_135 & 0.496\,/\,0.785\,/\,25 & \underline{0.502}\,/\,\underline{0.788}\,/\,25 & 0.483\,/\,0.775\,/\,24 & 0.499\,/\,0.787\,/\,\underline{25} & 0.487\,/\,0.808\,/\,\underline{25} & \textbf{0.559}\,/\,\textbf{0.823}\,/\,\textbf{27} \\
  GASA-02-02\_315 & \textbf{0.564}\,/\,\underline{0.779}\,/\,\textbf{26} & \underline{0.558}\,/\,0.775\,/\,\underline{26} & 0.466\,/\,0.745\,/\,23 & 0.450\,/\,0.738\,/\,22 & 0.500\,/\,0.767\,/\,24 & 0.501\,/\,\textbf{0.789}\,/\,24 \\
  \hline

  GASA-03\_105 & 0.507\,/\,0.788\,/\,25 & 0.508\,/\,\underline{0.790}\,/\,25 & 0.484\,/\,0.779\,/\,24 & \underline{0.523}\,/\,0.779\,/\,\underline{25} & 0.497\,/\,\textbf{0.801}\,/\,\underline{25} & \textbf{0.559}\,/\,\textbf{0.814}\,/\,\textbf{27} \\
  GASA-03\_285 & \underline{0.532}\,/\,\underline{0.820}\,/\,\underline{26} & 0.524\,/\,\textbf{0.821}\,/\,\underline{26} & 0.417\,/\,0.760\,/\,22 & 0.498\,/\,0.770\,/\,24 & 0.463\,/\,0.796\,/\,24 & \textbf{0.580}\,/\,0.806\,/\,\textbf{27} \\
  \hline

  GASA-04\_135 & 0.555\,/\,0.760\,/\,26 & 0.567\,/\,0.763\,/\,26 & 0.555\,/\,0.781\,/\,26 & \underline{0.602}\,/\,\underline{0.791}\,/\,\underline{27} & 0.586\,/\,\textbf{0.812}\,/\,\underline{27} & \textbf{0.628}\,/\,\textbf{0.813}\,/\,\textbf{28} \\
  GASA-04\_315 & 0.585\,/\,0.745\,/\,26 & 0.586\,/\,0.746\,/\,26 & 0.593\,/\,0.762\,/\,27 & \underline{0.640}\,/\,0.763\,/\,\underline{28} & 0.638\,/\,\textbf{0.796}\,/\,\underline{28} & \textbf{0.682}\,/\,\underline{0.791}\,/\,\textbf{29} \\
  \hline

GASA-05-O\_135 & 0.618\,/\,0.768\,/\,\textbf{27} & 0.603\,/\,0.768\,/\,\textbf{27} & 0.571\,/\,0.741\,/\,\underline{26} & \underline{0.620}\,/\,0.763\,/\,\textbf{27} & 0.596\,/\,\underline{0.770}\,/\,\textbf{27} & \textbf{0.628}\,/\,\textbf{0.778}\,/\,\textbf{27} \\
GASA-05-O\_315 & 0.614\,/\,0.764\,/\,27 & 0.609\,/\,0.765\,/\,27 & 0.626\,/\,0.782\,/\,\underline{28} & \underline{0.651}\,/\,0.788\,/\,\underline{28} & 0.639\,/\,\underline{0.796}\,/\,\underline{28} & \textbf{0.670}\,/\,\textbf{0.797}\,/\,\textbf{29} \\
GASA-05-W\_135 & 0.627\,/\,0.772\,/\,28 & 0.621\,/\,0.772\,/\,27 & 0.629\,/\,0.778\,/\,28 & \underline{0.671}\,/\,0.788\,/\,\underline{29} & 0.646\,/\,\underline{0.806}\,/\,\underline{29} & \textbf{0.685}\,/\,\textbf{0.808}\,/\,\textbf{30} \\
GASA-05-W\_315 & 0.598\,/\,\underline{0.772}\,/\,27 & 0.594\,/\,0.768\,/\,27 & 0.586\,/\,0.744\,/\,26 & \underline{0.637}\,/\,0.760\,/\,\underline{28} & 0.597\,/\,0.764\,/\,27 & \textbf{0.661}\,/\,\textbf{0.785}\,/\,\textbf{29} \\
\hline
GASA-06-B\_275 & 0.486\,/\,\underline{0.812}\,/\,\underline{25} & \textbf{0.505}\,/\,\textbf{0.815}\,/\,\textbf{26} & 0.267\,/\,0.709\,/\,15 & 0.330\,/\,0.787\,/\,19 & \underline{0.494}\,/\,0.775\,/\,23 & 0.479\,/\,0.796\,/\,22 \\
GASA-06-B\_95 & 0.303\,/\,\underline{0.838}\,/\,20 & 0.324\,/\,\textbf{0.841}\,/\,\underline{21} & 0.254\,/\,0.670\,/\,14 & 0.292\,/\,0.789\,/\,17 & \textbf{0.450}\,/\,0.796\,/\,\textbf{22} & \underline{0.394}\,/\,0.811\,/\,20 \\
\hline

  \end{tabularx}
\end{table*}

\subsubsection{Temporal Stability Metrics}
The temporal stability metrics quantify how far into the future a model can produce stable and reliable forecasts. In this study, we define Stable Lead Time, which is quantified as the length of the forecasting horizon, beginning at $t=1$, over which the groud-truth remain within the interval. Let $L_i$ denote the length, formally,
\begin{equation}
L_i
=
\max
\left\{
k \in \{0,\dots,40\}
\;\middle|\;
h_{i,t} = 1,\ \forall t \in \{1,\dots,k\}
\right\}.
\end{equation}
If $h_{i,1}=0$, $L_i=0$. The Stable Lead Time over all target timesteps $i$ is computed as
\begin{equation}
\text{SLT}
=
\frac{1}{N}
\sum_{i=1}^{N}
L_i.
\end{equation}
\subsubsection{Predictive Uncertainty}
\paragraph{Normalized Mean Prediction Interval Width (NMPIW)}
 
We utilize NMPIW \cite{khosravi2011comprehensive} to quantify how wide the prediction intervals are on average. Intuitively, a wider interval reflects greater uncertainty and lower precision. To ensure comparability across sensors and locations, we normalize the average prediction interval width by the historical min - max range of observations for the corresponding sensor/location:

\begin{equation}
\text{NMPIW}
=
\frac{1}{N}
\sum_{i=1}^{N}
\left(
\frac{
\frac{1}{40}
\sum_{t=1}^{40}
\bigl(\hat{u}_{i,t} - \hat{\ell}_{i,t}\bigr)
}{
\max(Y_{hist}) - \min(Y_{hist})
}
\right),
\end{equation}
where $Y_{hist}$ denotes the historical pedestrian flow count observations for corresponding tested sensor.

\begin{table*}[htbp]
  \centering
  \small
  \setlength{\tabcolsep}{1.5pt}
  \renewcommand{\arraystretch}{1.1}
  \caption{Normalized Mean Prediction Interval Width (NMPIW) Analysis: Core Operational Hours \textbf{08:00 - 22:00}. Bold: Best, Underline: Second Best. Fixed: fixed context window, Increasing: increasing context window.}

  \newcolumntype{C}[1]{>{\centering\arraybackslash}m{#1}}
  \label{tab:NMPIW}
  \begin{tabularx}{\textwidth}{|C{3.2cm}|*{6}{>{\centering\arraybackslash}X|}}
  \hline

  \multicolumn{1}{|c|}{\textbf{Sensor ID}} &
  \multicolumn{1}{c|}{\makecell[c]{\textbf{TimesFM}\\\textbf{(Fixed)}}} &
  \multicolumn{1}{c|}{\makecell[c]{\textbf{TimesFM}\\\textbf{(Increasing)}}} &
  \multicolumn{1}{c|}{\makecell[c]{\textbf{Chronos-2}\\\textbf{(Fixed)}}} &
  \multicolumn{1}{c|}{\makecell[c]{\textbf{Chronos-2}\\\textbf{(Increasing)}}} &
  \multicolumn{1}{c|}{\makecell[c]{\textbf{Chronos-2}\\\textbf{(Fixed}\\\textbf{+ TimeFeature)}}} &
  \multicolumn{1}{c|}{\makecell[c]{\textbf{Chronos-2}\\\textbf{(Increasing}\\\textbf{+ TimeFeature)}}} \\
  \hline

  GASA-01-A1\_135 & \underline{0.166} & \textbf{0.166} & 0.170 & 0.172 & 0.172 & 0.178 \\
  
  GASA-01-A1\_315 & \underline{0.255} & \textbf{0.254} & 0.260 & 0.260 & 0.263 & 0.270 \\
  
  GASA-01-A2\_135 & 0.294 & 0.291 & \underline{0.283} & \textbf{0.282} & 0.292 & 0.298 \\
  
  GASA-01-A2\_315 & 0.211 & 0.210 & \underline{0.205} & \textbf{0.201} & 0.206 & 0.213 \\
 
  GASA-01-B\_135 & \underline{0.177} & \textbf{0.175} & 0.184 & 0.189 & 0.183 & 0.190 \\
  
  GASA-01-B\_315 & 0.159 & 0.157 & \underline{0.154} & \textbf{0.150} & 0.154 & 0.161 \\
  
  GASA-01-C\_135 & \underline{0.234} & \textbf{0.233} & 0.258 & 0.258 & 0.256 & 0.261 \\
 
  GASA-01-C\_315 & \underline{0.169} & \textbf{0.164} & 0.181 & 0.178 & 0.179 & 0.185 \\
  \hline
  GASA-02-01\_135 & \underline{0.109} & \textbf{0.107} & 0.116 & 0.114 & 0.118 & 0.121 \\
  
  GASA-02-01\_315 & 0.197 & 0.195 & \underline{0.179} & \textbf{0.169} & 0.180 & 0.188 \\
  
  GASA-02-02\_135 & 0.259 & 0.257 & \underline{0.236} & \textbf{0.233} & 0.238 & 0.250 \\
  
  GASA-02-02\_315 & \underline{0.104} & \textbf{0.103} & 0.121 & 0.121 & 0.123 & 0.132 \\
  \hline
  GASA-03\_105 & 0.283 & 0.284 & \textbf{0.244} & \underline{0.252} & 0.261 & 0.274 \\
  
  GASA-03\_285 & 0.224 & 0.221 & \underline{0.187} & \textbf{0.184} & 0.195 & 0.202 \\
  \hline
  GASA-04\_135 & 0.245 & 0.249 & \underline{0.225} & \textbf{0.220} & 0.237 & 0.232 \\
  
  GASA-04\_315 & 0.218 & 0.217 & \underline{0.208} & \textbf{0.194} & 0.215 & 0.210 \\
  \hline
  GASA-05-O\_135 & 0.137 & 0.136 & \textbf{0.129} & 0.134 & \underline{0.133} & 0.137 \\
  
  GASA-05-O\_315 & 0.132 & 0.131 & \underline{0.131} & 0.133 & \textbf{0.130} & 0.135 \\
  
  GASA-05-W\_135 & 0.100 & \textbf{0.099} & 0.101 & 0.102 & \underline{0.099} & 0.105 \\
  
  GASA-05-W\_315 & \underline{0.113} & \textbf{0.113} & 0.115 & 0.119 & 0.117 & 0.122 \\
  \hline
  
  GASA-06-B\_95 & \textbf{0.077} & \underline{0.078} & 0.157 & 0.149 & 0.156 & 0.144 \\
  
  GASA-06-B\_275 & \underline{0.117} & \textbf{0.117} & 0.182 & 0.176 & 0.186 & 0.179 \\
  \hline

  \end{tabularx}
\end{table*}

\paragraph{Scaled Winkler Score (SWS)}

To provide a unified assessment that complements the reliability, stability, and uncertainty metrics, we employ the Scaled Winkler Score (SWS) \cite{winkler1972decision} as an overall measure of probabilistic forecast quality. SWS is an interval scoring rule that balances two desirable properties of interval forecasts: reliable coverage of the ground truth and sharpness (i.e., avoiding unnecessarily wide intervals).

For a target coverage level of $(1-\alpha)\times100\%$ (with $\alpha = 0.2$ corresponding to the 10th - 90th percentile interval), the step-wise Winkler score $S_{i,t}$ at lead step $t$ is defined as:
\begin{equation}
S_{i,t} =
\begin{cases}
\hat{u}_{i,t} - \hat{\ell}_{i,t},
& \text{if } \hat{\ell}_{i,t} \le y_{i,t} \le \hat{u}_{i,t}, \\[6pt]
(\hat{u}_{i,t} - \hat{\ell}_{i,t}) + \dfrac{2}{\alpha}\,(\hat{\ell}_{i,t} - y_{i,t}),
& \text{if } y_{i,t} < \hat{\ell}_{i,t}, \\[6pt]
(\hat{u}_{i,t} - \hat{\ell}_{i,t}) + \dfrac{2}{\alpha}\,(y_{i,t} - \hat{u}_{i,t}),
& \text{if } y_{i,t} > \hat{u}_{i,t}.
\end{cases}
\end{equation}

If the interval covers the ground truth, the score reduces to the interval width; valid intervals with lower predictive uncertainty have lower scores. If coverage fails, the score adds an extra penalty scaled by the coefficient $2/\alpha$ (e.g., $2/\alpha=10$ for an 80\% interval with $\alpha=0.2$). The penalty increases linearly with the distance between $y_{i,t}$ and the nearest interval bound, which penalizes overconfident intervals that miss the target by a large margin.

We normalize the time-averaged Winkler score by the historical min - max range of the corresponding sensor/location to ensure fair comparison across sensors with heterogeneous pedestrian flow scales. The Scaled Winkler Score is computed as:
\begin{equation}
\text{SWS}
=
\frac{1}{N}
\sum_{i=1}^{N}
\left(
\frac{
\frac{1}{40}\sum_{t=1}^{40} S_{i,t}
}{
\max(Y_{hist}) - \min(Y_{hist})
}
\right),
\end{equation}
A lower SWS indicates superior probabilistic forecast quality, reflecting a favorable balance between narrow prediction intervals and reliable coverage.

\begin{figure}[]
    \centering
    \includegraphics[width=\linewidth]{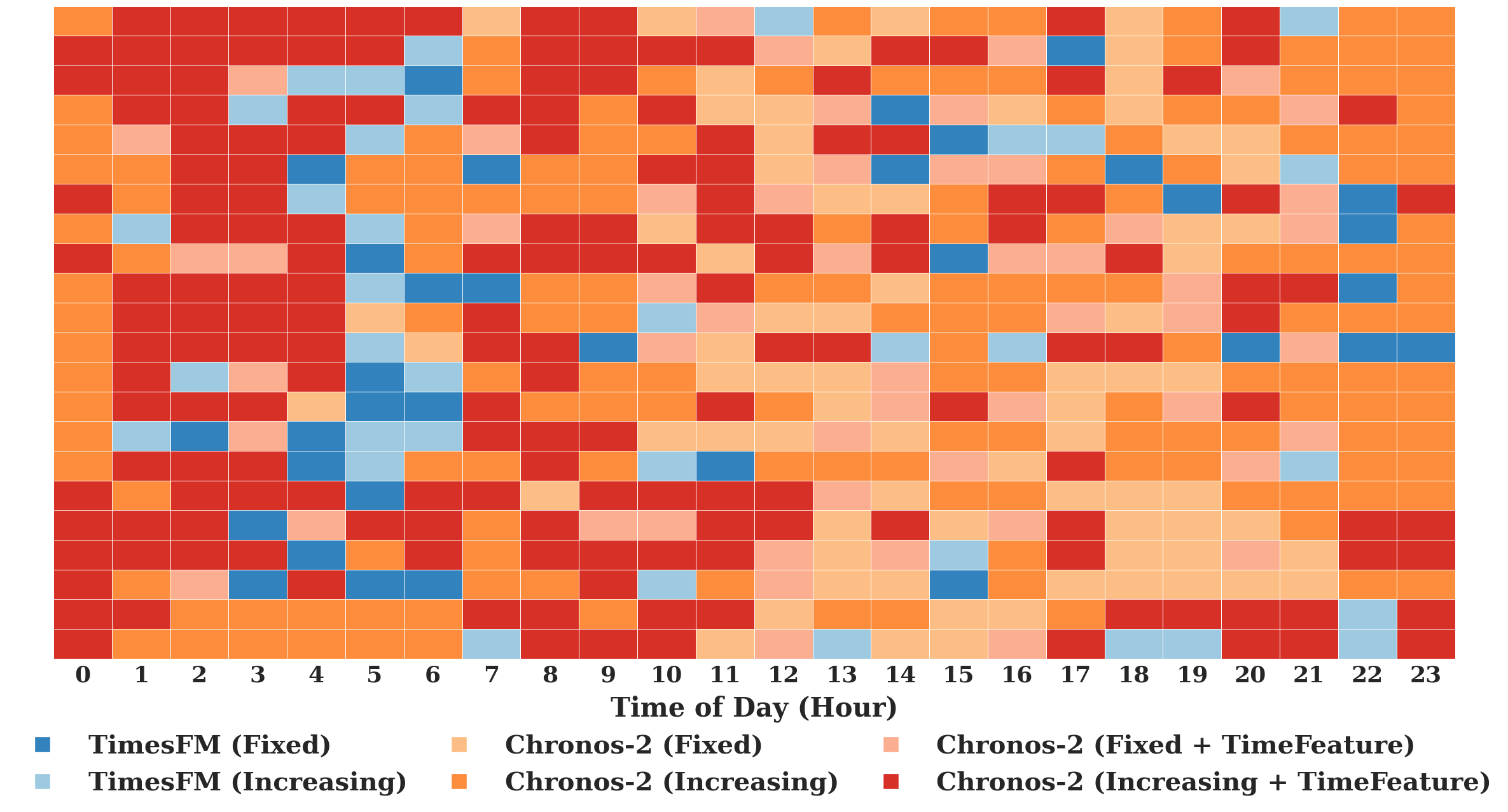}
    \caption{Best-performing model per sensor and time of day, defined by the minimum Scaled Winkler Score. Each cell indicates the best model for the corresponding sensor and hour pair.}
    \label{fig:Winner_Model_Heatmap_24H}
\end{figure}

\section{Results}
\label{sec:Results}
Table~\ref{tab:over_all} reports model performance during the core operational hours (08{:}00 - 22{:}00) for all test days. Chronos-2 (Increasing + TimeFeature) ranks best on nearly all sensors in FHC, NHC, and SLT. More specifically, the results show that zero-shot foundation models provide stable prediction intervals under 90-second updates: across almost all sensors, SLT reaches about 20 steps ahead (30 minutes), and Chronos-2 (Increasing + TimeFeature) reaches about 27 to 30 steps ahead (40 to 45 minutes). Operational short-term forecasting in Intelligent Transportation Systems is commonly evaluated at 15-30 minute horizons \cite{muhammed2024gat} \cite{saleh2023metro}. For crowd management, a lead time of 30 minutes provide crowd managers with a favorable window to implement appropriate response measures in many scenarios \cite{li2019short} \cite{liu2023bi}. %In summary, the SLT results indicate that zero-shot foundation models provide necessary lead time for proactive interventions.

In addition, across most sensors, Chronos-2 model with increasing input context and additional temporal features outperforms the ones without. This pattern suggests that prolonged input length and additional covariates improve probabilistic forecasting in this setting. In contrast, TimesFM shows limited or no gains from the increasing context window. 

Although Chronos2-based models lead overall, TimesFM (Increasing) performs best on GASA-06-B\_275 and ranks first across all three metrics. This sensor has low pedestrian flow counts and limited variability over the 5-day period, with a clear peak only on the final day. The night-time experiments (22{:}00 - 08{:}00) across all sensors also show similar results: TimesFM is more reliable in low-flow periods; due to page limitations, these results are not shown here.

Table~\ref{tab:NMPIW} reports the NMPIW comparison during the core operational hours (08{:}00 - 22{:}00). Smaller NMPIW corresponds to tighter intervals and higher confidence. Without time covariates, the increasing input models show lower NMPIW values than the fixed input models on many sensors. This suggests that additional context can reduce interval width in this rolling forecasting setup. Meanwhile, we observe clear sensor-dependent differences in NMPIW: low-variability sensors (eg. GASA-05-W\_135) tend to show tighter intervals, while high-variability sensors (eg. GASA-01-A2\_135) with more frequent sharp surges exhibit wider intervals. In addition, for Chronos2, the TimeFeature variants show larger NMPIW on several sensors compared with their counterparts without time covariates. This pattern indicates higher uncertainty or lower confidence after adding time covariates, a plausible explanation is that time features help low-variability sensors with stable patterns, reducing uncertainty and predictive interval. For high-variability sensors with strong fluctuations and more complex data distributions, simple time covariates are less informative and can add extra noise, so the model becomes less confident and outputs tighter intervals to maintain coverage.

Fig.~\ref{fig:Winner_Model_Heatmap_24H} reports, for each sensor and hour, the model with the lowest Scaled Winkler Score (SWS). Chronos-2-based models appear most frequently as the best choice across sensors, with Chronos-2 (Increasing + TimeFeature) ranking the best overall.  Specifically, we observe a time-of-day pattern with the sensor in the figure. In the early-morning low-flow hours (around 04:00 - 06:00), TimesFM appears more frequently as the SWS-best model for several sensors, while during daytime and core operational hours, the best choices are dominated by Chronos-2. In addition, Chronos-2 (Increasing + TimeFeature) typically gets the best SWS performance at night and in the early morning (19:00–04:00), when the within-day pattern is more regular.

\subsection{Case Study}

To supplement the overall results, we conducted three individual sensor case studies to examine the operational reliability under different data distributions. Case I examines the performance of the model under sustained high-level and rapid changes in pedestrian flow. Case II focuses on sudden surges and pattern shifts. Case III assesses the model's sensitivity to corrupted or missing data, and examines the robustness of uncertainty estimation.

\begin{figure}[htbp]
    \centering
    \includegraphics[width=\linewidth]{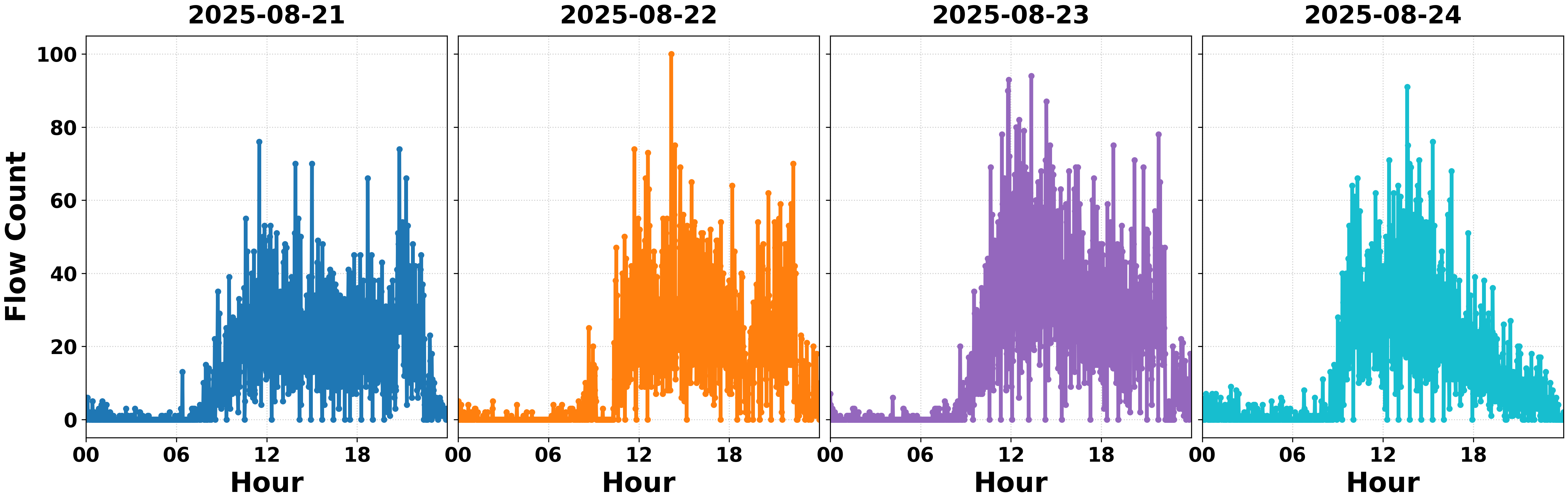}
    \caption{The ground-truth pedestrian flow counts recorded by GASA-01-A2\_135 over the four-day prediction period.}
    \label{fig:case1_volatility}
\end{figure}

\subsubsection{Case I: High Volatility and Rapid Fluctuations} \label{sec:case_volatility} 

As shown in Fig.~\ref{fig:case1_volatility}, in this case (Sensor ID: \texttt{GASA-01-A2\_135}), the pedestrian flow reaches relatively high levels and shows high volatility, with rapid fluctuations over short intervals.

Based on Table~\ref{tab:case1_volatility}, Chronos-based models show stronger overall performance in this high-volatility case. The Increasing setting improves Chronos over the Fixed setting on both days. In addition, the Increasing + TimeFeature setting provides only marginal improvements in NHC and SLT for Chronos on both days. These results indicate limited gains from time covariates for this case definition, even though they increase coverage-related metrics. Moreover, TimesFM is less competitive compare to Chronos-2 in this high-volatility case.  One possible explanation is that Chronos-2 benefits more from longer history due to its modeling formulation. Chronos-2 performs generative modeling on discretized series, and its training objective encourages the use of extended context to constrain the conditional distribution of future values. As a result, information in a long history (e.g., recurring patterns, amplitude ranges, and volatility scales) is easier to integrate into regime inference. In contrast, TimesFM relies on patch-based representations that emphasize local shape matching. Under a short event window and strong non-stationarity, additional history may translate less directly into stable gains and can introduce extra noise.

\begin{table}[t]
  \centering
  \small
  \setlength{\tabcolsep}{3pt}
  \renewcommand{\arraystretch}{1.05}
  \caption{Case Study Performance Comparison (Core Hours 08:00 - 22:00). 
  Metrics: NHC, SLT, NMPIW, SWS.
  Each cell shows \textbf{2025-08-23} (top) and \textbf{2025-08-24} (bottom).
  \textbf{Bold}: 1st, \underline{Underline}: 2nd (ties are marked).}
  \label{tab:case1_volatility}

  \begin{tabular}{|l|cccc|}
    \hline
    \textbf{Model} & \textbf{NHC} & \textbf{SLT} & \textbf{NMPIW} & \textbf{SWS} \\
    \hline
    TimesFM (Fixed) &
    \makecell{0.764\\\underline{0.775}} &
    \makecell{26\\26} &
    \makecell{0.682\\\underline{0.503}} &
    \makecell{1.029\\0.791} \\
    \hline
    TimesFM (Increasing) &
    \makecell{0.759\\0.769} &
    \makecell{25\\26} &
    \makecell{0.671\\\underline{0.503}} &
    \makecell{1.030\\0.779} \\
    \hline
    Chronos (Fixed) &
    \makecell{0.745\\0.733} &
    \makecell{23\\25} &
    \makecell{\underline{0.636}\\\textbf{0.497}} &
    \makecell{1.006\\0.745} \\
    \hline
    Chronos (Increasing) &
    \makecell{0.784\\\underline{0.775}} &
    \makecell{\underline{27}\\\underline{27}} &
    \makecell{\textbf{0.632}\\0.517} &
    \makecell{\textbf{0.945}\\\textbf{0.696}} \\
    \hline
    Chronos (Fixed + TimeFeature) &
    \makecell{\underline{0.802}\\0.773} &
    \makecell{26\\26} &
    \makecell{0.663\\0.510} &
    \makecell{0.998\\0.742} \\
    \hline
    Chronos (Increasing + TimeFeature) &
    \makecell{\textbf{0.819}\\\textbf{0.786}} &
    \makecell{\textbf{28}\\\textbf{28}} &
    \makecell{0.666\\0.536} &
    \makecell{\underline{0.949}\\\underline{0.698}} \\
    \hline
  \end{tabular}
\end{table}

\subsubsection{Case II: Sudden Spike and Pattern Shift}
\begin{figure}[htbp]
    \centering
    \includegraphics[width=\linewidth]{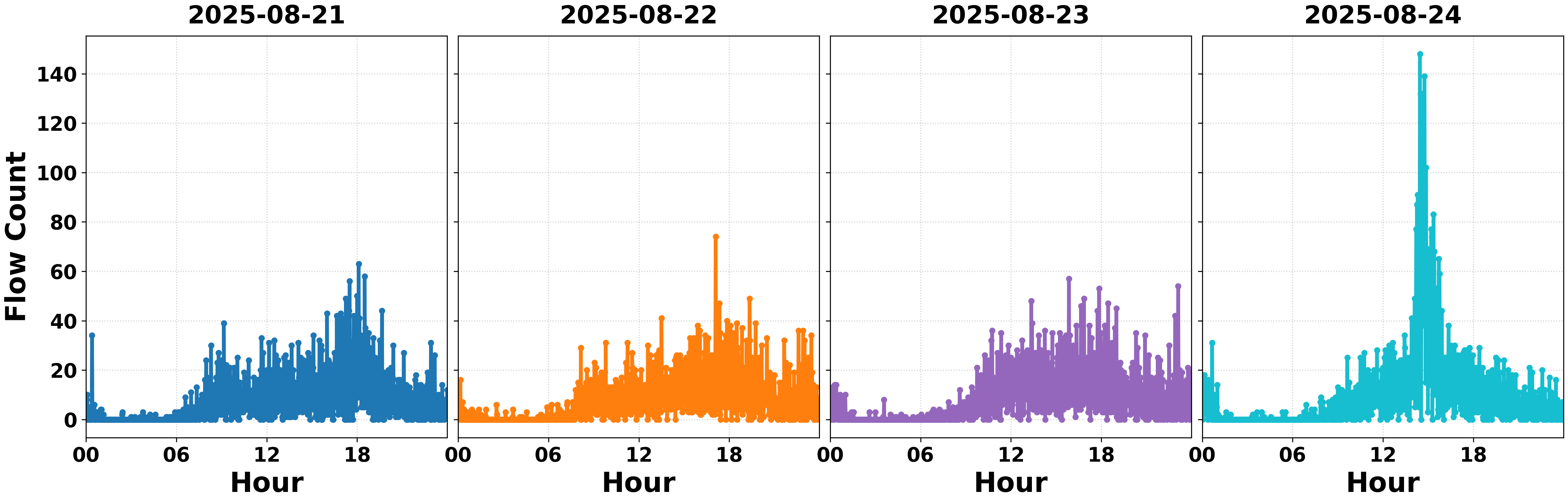}
    \caption{The ground-truth pedestrian flow counts recorded by GASA-05-O\_135 over the four-day prediction period}
    \label{fig:case2_gt}
\end{figure}

Fig.~\ref{fig:case2_gt} shows a clear pattern shift for sensor \texttt{GASA-05-O\_135}: the first three days follow a low and relatively stable daily pattern with moderate peaks (approximately 38 - 42), while the final day (2025-08-24) exhibits a sudden peak around 14:00 - 16:00, reaching a maximum of 148.

\begin{figure}[tbp]
    \centering
    \includegraphics[width=\linewidth]{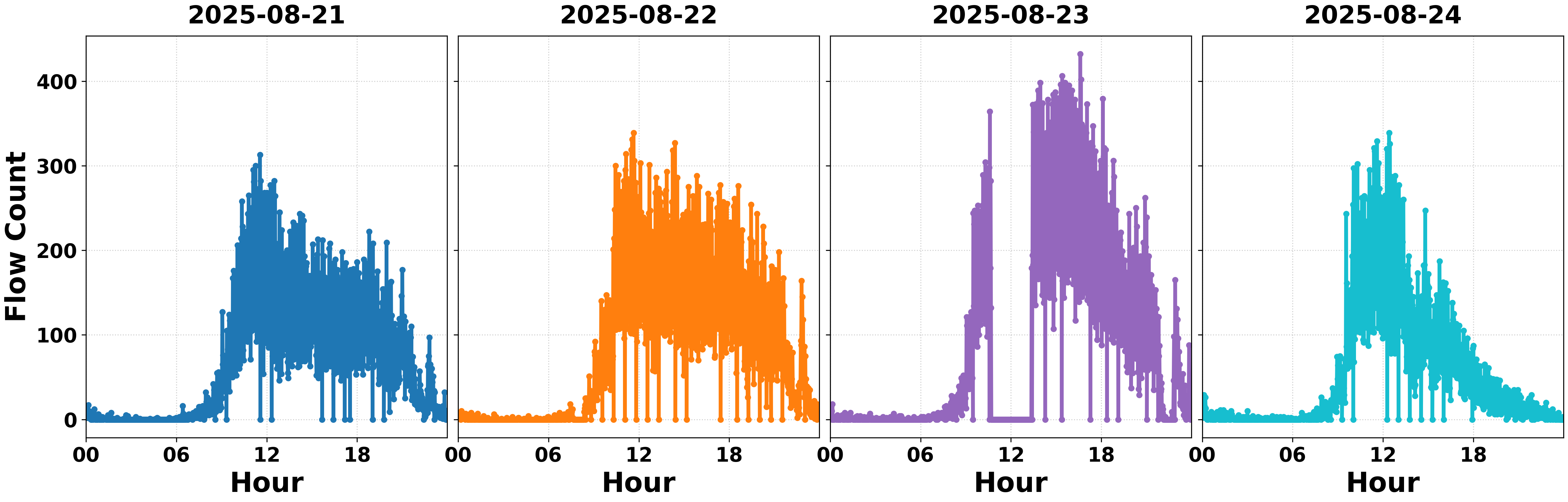}
    \caption{The ground-truth pedestrian flow counts recorded by GASA-02-02\_135 over the four-day prediction period}
    \label{fig:case3}
\end{figure}

Table~\ref{tab:case2_metrics} reports the SLT and NHC results for all models at the sudden peak window. Chronos (Increasing + TimeFeature) achieves the best performance on both metrics, and Chronos (Increasing) ranks second, which indicates that Chronos benefits from longer context and further improves when time covariate are available in this case. A potential explanation for the lower performance of TimesFM in the sudden spike window relates to representation bias under distribution shift. Patch-based Transformers compress the input into local patch tokens to capture short-term shapes efficiently \cite{nie2022time}, but recent work shows that over-reliance on patching can hurt generalization when local structure changes abruptly \cite{Luo2024DeformableTST}, and this sensitivity increases under patch-level distribution shift, where patterns drift across segments \cite{sun2024learning}.

\begin{table}[!htbp]
  \centering
  \small
  \caption{Response analysis during the surge window (14:00 - 16:00) on 2025-08-24. SLT reflects how many steps in advance the model reliably predicts the target value, while NHC indicates one-step-ahead reliability.}
  \label{tab:case2_metrics}
  \begin{tabular}{l cc}
    \toprule
    \textbf{Model} & \textbf{SLT} & \textbf{NHC} \\
    \midrule
    TimesFM (Fixed) & 14 & 0.625 \\
    TimesFM (Increasing) & 12 & 0.600 \\
    Chronos (Fixed) & 13 & 0.700 \\
    Chronos (Increasing) & \underline{16} & \underline{0.725} \\
    Chronos (Fixed + TimeFeature) & 15 & 0.662 \\
    Chronos (Increasing + TimeFeature) & \textbf{17} & \textbf{0.738} \\
    \bottomrule
  \end{tabular}
\end{table}

\subsubsection{Case III: Robustness to Missing Data}

In this case (Sensor ID: \texttt{GASA-02-02\_135}), we analyze model performance when approximately 30\% of the historical data during 2025-08-23 11{:}30 - 13{:}30 is missing (represented as zeros), as shown in Fig.~\ref{fig:case3}.

\begin{figure}[htbp]
    \centering
    \includegraphics[width=\linewidth]{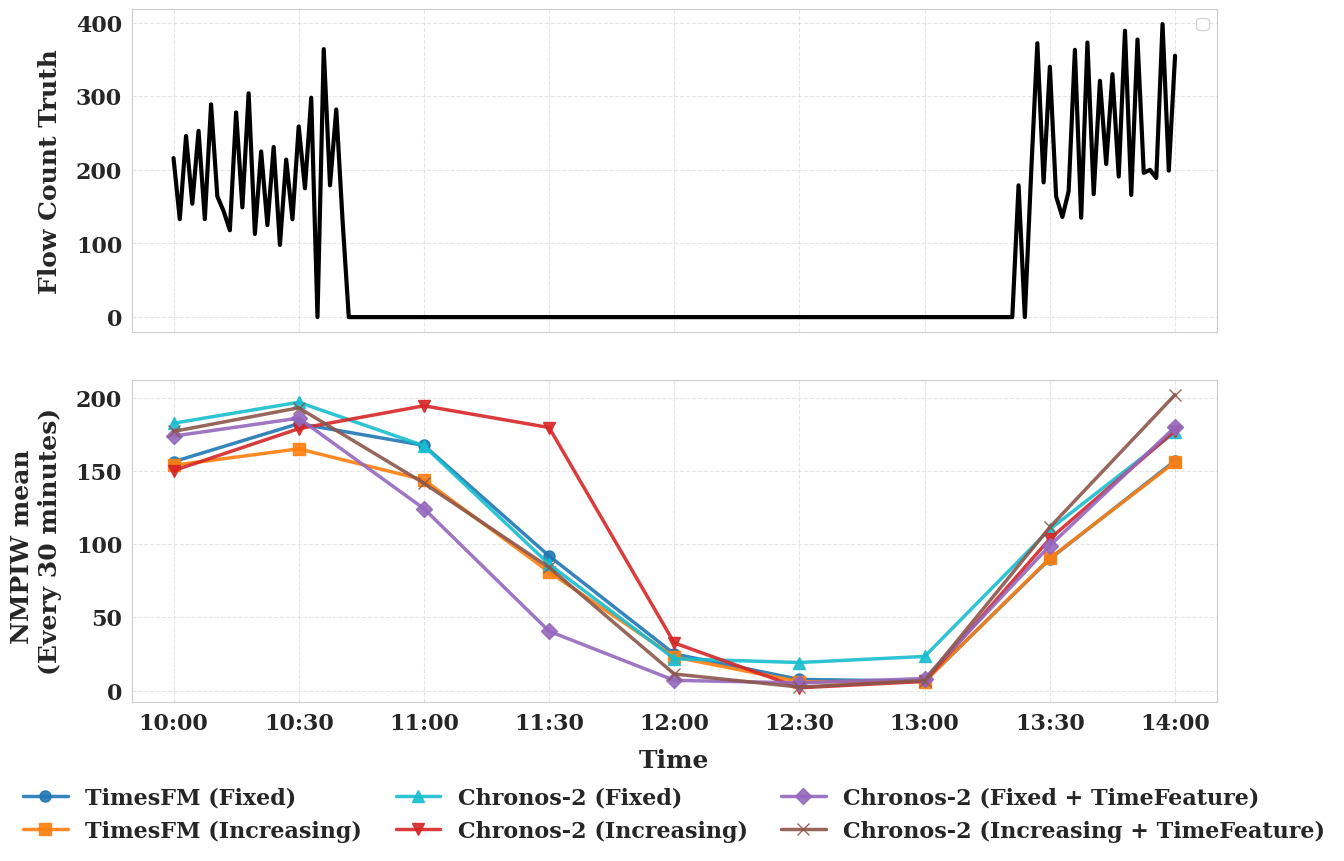}
    \caption{Case III: Uncertainty on Data Missing (2025-08-23)}
    \label{fig:case3_gt_1}
\end{figure}

% Under this adversarial condition, robustness is defined not just by accuracy, but by reliable uncertainty quantification. As shown in Figure \ref{fig:case3}, \textbf{Chronos} responds to missing data by significantly widening its prediction intervals (high Relative Width), effectively communicating its uncertainty to the operator. This "fail-safe" behavior is preferable in safety-critical applications compared to models that produce narrow, confident, but erroneous predictions.

% \begin{figure}[htbp]
%     \centering
%     \includegraphics[width=\linewidth]{Fig/Case3.png}
%     \caption{Case III: Next Day Uncertainty After Data Missing (2025-08-24)}
%     \label{fig:case3_gt_2}
% \end{figure}

As shown in Fig.~\ref{fig:case3_gt_1}, the missing data segment induces distinct uncertainty dynamics across models. Chronos-2 (Increasing) responds to missing data by significantly widening its prediction intervals (higher NMPIW) at the onset of the missing data, effectively communicating its uncertainty to the operator. This ``fail-safe" behavior is preferable in safety-critical applications compared to models that produce narrower, over-confident, but erroneous predictions.

% widens its prediction intervals at the onset of the data missing occurrence, and then the intervals gradually tighten as the zero value continues to dominate the input context. A plausible explanation is that, with a long context, the sudden onset of a missing data segment contradicts the earlier history, thereby the model becomes less confident and expresses higher uncertainty through wider intervals. From an operational perspective, the NMPIW from Chronos-2 (Increasing) around the onset and recovery of missing data segments can provide insights for crowd managers, which can serve as a signal of missing data and sensor failure. More broadly, this uncertainty response can be viewed as a robust fail-safe mechanism.

\begin{figure}[t]
    \centering
    \includegraphics[width=\linewidth]{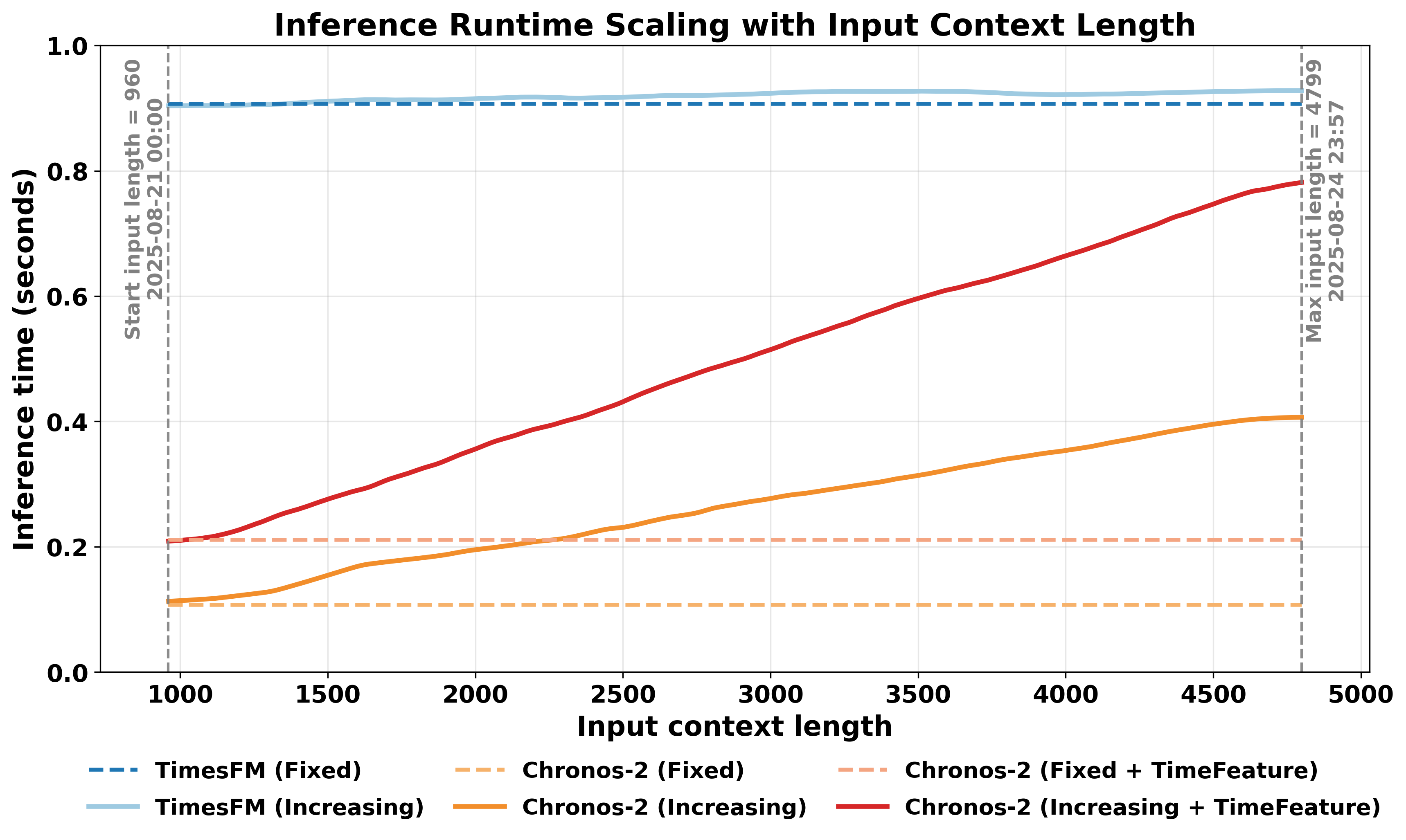}
    \caption{CPU inference runtime scaling with input context length. Dashed horizontal lines indicate fixed-context inference using 960 historical observations, while solid curves indicate increasing-context inference where the input context grows from 960 to 4799 observations. All models forecast 40 future steps.}
    \label{fig:runtime_scaling}
\end{figure}

\begin{table*}[t]
\centering
\scriptsize
\caption{Model checkpoint size and CPU resident set size memory footprint during model startup, warm-up inference, and rolling inference.}
\label{tab:rss_memory_footprint}
\resizebox{\textwidth}{!}{%
\begin{tabular}{l|rrrr}
\hline
Model setting
& Checkpoint size
& Startup peak RSS
& Warm-up peak RSS
& Rolling peak RSS \\
& (MB) & (MB) & (MB) & (MB) \\
\hline
TimesFM (Fixed) & 925 & 886.96 & 94.07 & 48.24 \\
TimesFM (Increasing) & 925 & 886.96 & 94.07 & 35.95 \\
Chronos-2 (Fixed) & 478 & 443.80 & 431.27 & 0.31 \\
Chronos-2 (Increasing) & 478 & 475.60 & 462.11 & 51.25 \\
Chronos-2 (Fixed + TimeFeature) & 478 & 474.14 & 461.32 & 5.30 \\
Chronos-2 (Increasing + TimeFeature) & 478 & 477.81 & 464.88 & 44.88 \\
\hline
\end{tabular}
}
\end{table*}

\subsection{Operational Memory Footprint}

To evaluate operational memory footprint, we measure process Resident Set Size (RSS) across three phases: model setup, warm-up inference and rolling inference. All reported RSS metrics are averaged over 22 directional sensor time series. Model setup refers to loading the foundation model for CPU inference. Warm-up inference denotes the first forecasting call after setup and is measured separately from subsequent rolling inference, following common inference benchmarking practice \cite{reddi2020mlperf}. We define startup peak RSS as the maximum RSS increase observed during model setup, before the first forecasting. Warm-up peak RSS is measured separately as the maximum RSS increase during the first forecasting call after setup. Rolling peak RSS is measured after warm-up over the full rolling forecasting period.

The results show that TimesFM reaches the highest startup memory peak, approximately 887 MB, whereas Chronos-2 remains within approximately 444--478 MB across all settings. This is consistent with the larger checkpoint size of TimesFM, with a checkpoint size of 925 MB compared with 478 MB for Chronos-2. However, the comparison also shows that runtime RSS behavior is not determined by checkpoint size alone. Chronos-2 shows a much larger warm-up peak RSS, indicating that substantial additional memory is allocated during the first inference call. After warm-up, all settings remain stable during rolling inference, with rolling peak RSS staying below 52 MB. From an operational perspective, all measured startup peak RSS values remain below 1 GB on the CPU-only experimental platform. These results suggest that, once the setup and warm-up phases are completed, the memory footprint remains stable during rolling inference under the evaluated setting.

\subsection{Computational Feasibility}

Computational feasibility is essential for real-time crowd-management deployment. Fig.~\ref{fig:runtime_scaling} reports the CPU inference runtime of the evaluated foundation models under all input strategies. The sensor data are updated every 90 seconds, while the inference runtime of all evaluated foundation models remains below one second across the tested context range, which indicates that the models can generate forecasts fast enough without GPU acceleration. Meanwhile, the models show different runtime scaling patterns.  Chronos-2 runtime increases more clearly as the input context grows, especially when TimeFeature covariates are included. This indicates that longer histories and additional temporal covariates introduce extra computational overhead for Chronos-2. In contrast, TimesFM has a higher inference runtime, but its runtime curve under the increasing strategy grows only slightly and remains close to that of the fixed strategy. This pattern is consistent with the patching design of TimesFM, where contiguous time steps are first grouped into input patches and then converted into tokens for the stacked Transformer layers; the effective token sequence grows more slowly than the raw number of observations.

\begin{figure}[t]
    \centering
    \includegraphics[width=0.95\linewidth]{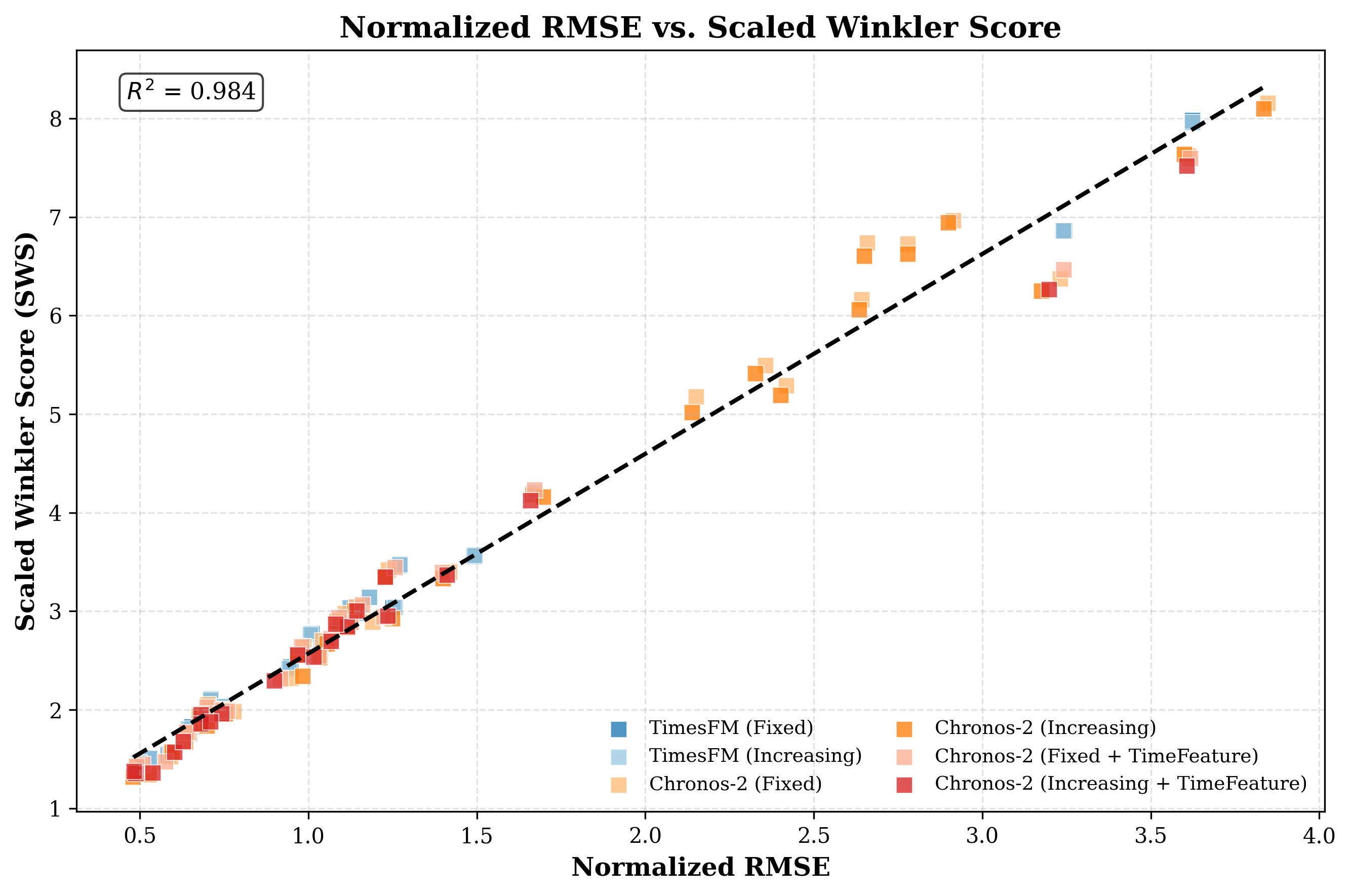}
    \caption{Relationship between normalized RMSE and Scaled Winkler Score across all evaluated model settings and 22 directional sensor time series. Each point represents one model setting evaluated on one time series.}
    \label{fig:rmse_sws_relation}
\end{figure}

\subsection{Accuracy Consistency of Scaled Winkler Score}

The main evaluation in this study focuses on interval-based reliability and probabilistic forecast quality. However, these metrics primarily characterize uncertainty representation, and are not intended to serve as direct point-accuracy measures. Therefore, we additionally examine whether SWS, as an interval-based probabilistic metric, remains consistent with conventional point-error behavior. We compare SWS with normalized RMSE, computed using the median forecast ($50^{\mathrm{th}}$ percentile), which is commonly used as the point forecast when quantile predictions are available. We then compare normalized RMSE with SWS across all evaluated model settings and 22 directional sensor time series. Each point in Fig.~\ref{fig:rmse_sws_relation} represents one aggregated evaluation result. More specifically, one point corresponds to one model setting on one directional sensor time series.

Fig.~\ref{fig:rmse_sws_relation} shows a strong positive linear relationship between normalized RMSE and SWS, with $R^2 = 0.984$. Model settings with larger point-error also tend to obtain worse SWS scores. This result shows that SWS is able to provide an accuracy-consistent measure of probabilistic forecast quality. Meanwhile, SWS contains information beyond RMSE because it also accounts for interval width and penalties for uncovered observations. Therefore, SWS can be used as a compact reliability-oriented metric, while more detailed evaluation can be obtained by combining it with coverage metrics, stable lead time, and interval-width measures such as NMPIW.

\FloatBarrier
\section{discussion and conclusion}

In this paper, the experimental results analyse the reliability of time-series foundation models for pedestrian-flow forecasting during special events. Using the SAIL2025 event, we show that these models can provide reliable multi-step probabilistic forecasts with a lead time of around 30 minutes, which is adequate for crowd management Among the evaluated variants, Chronos-2 performs better than TimesFM in most cases, making Chronos-2 the better choice for deployment in this special event setting. Longer context windows lead to a clear and consistent improvement for Chronos-2. In contrast, adding simple time features (date/hour) as the covariate brings only limited improvements. 

The paper also provide decision-oriented reliability evaluation metrics, which could help crowd managers select the most suitable models, including coverage, stable lead time, and predictive uncertainty. In future work, we plan to add more data to the forecasting, such as other sensor data and public data sources (e.g., weather and event information), and test whether more contextual data can further improve the performance.

%%%%%%%%%%%%%%%%%%%%%%%%%%%%%%%%%%%%%%%%%%%%%%%%%%%%%%%%%%%%%%%%%%
\section*{ACKNOWLEDGMENTS}
This work was supported by the AI-COMPASS project, funded by the Dutch Research Council (NWO) under grant number KICH1.VE04.22.007.

%%%%%%%%%%%%%%%%%%%%%%%%%%%%%%%%%%%%%%%%%%%%%%%%%%%%%%%%%%%%%%%%%%
%\addtolength{\textheight}{-12cm}
%\vspace{10mm}
\bibliographystyle{IEEEtran}
% Your .bib file here
\bibliography{root} 

\end{document}